\documentclass{article} 
\usepackage{colm2025_conference}

\usepackage{microtype}
\usepackage{hyperref}
\usepackage{url}
\usepackage{booktabs}
\usepackage{multirow}
\usepackage[table]{xcolor}
\usepackage[utf8]{inputenc}
\usepackage{lineno}
\usepackage{xcolor}
\usepackage{titlesec}
\usepackage[most]{tcolorbox}
\usepackage{caption}
\usepackage{wrapfig} 
\definecolor{darkblue}{rgb}{0, 0, 0.5}
\hypersetup{colorlinks=true, citecolor=darkblue, linkcolor=darkblue, urlcolor=darkblue}
\usepackage{graphicx}
\usepackage{subcaption}
\usepackage{colortbl}
\usepackage{amsmath}
\usepackage{siunitx}    
\usepackage{amssymb} 
\usepackage{hyperref}
\usepackage{url}
\usepackage{graphicx}
\usepackage{wrapfig} 
\usepackage{mathrsfs}
\usepackage{booktabs}
\usepackage{pifont} 
\usepackage[table]{xcolor}
\usepackage{graphicx}
\usepackage{subcaption}
\usepackage{tikz}
\usepackage{eso-pic} 
\usepackage{multirow}

\definecolor{richpurple}{RGB}{75,46,131}
\newcommand{\purplecomic}[1]{%
  {\color{richpurple}\selectfont #1}%
}

\newcommand{\blackcomic}[1]{%
  {\color{black}\selectfont #1}%
}

\newcommand\eat[1]{}

\title{
  \purplecomic{\includegraphics[height=24pt]{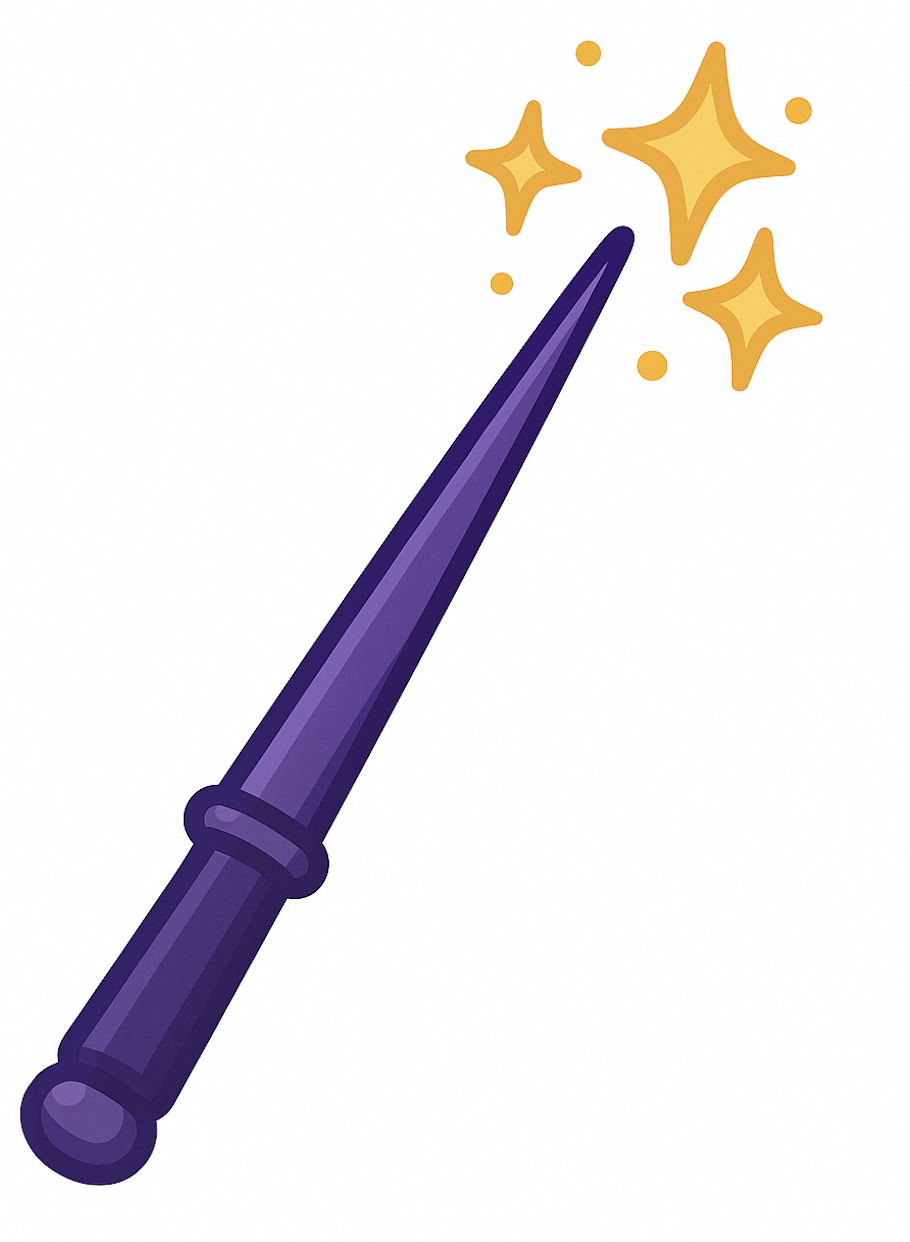} \textbf{CuES}: 
  \blackcomic{A Curiosity-driven and Environment-grounded\\ Synthesis Framework for Agentic RL}
}
}

\author{%
  Shinji~Mai$^{*}$\quad
  Yunpeng~Zhai$^{*}$\quad
  Ziqian~Chen\quad
  Cheng~Chen\quad
  Anni~Zou\quad
  Shuchang~Tao\quad
  Zhaoyang~Liu$^{\dagger}$\quad
  Bolin~Ding$^{\dagger}$\quad
  \\
  [0.5em]
   Tongyi Lab \includegraphics[height=14pt]{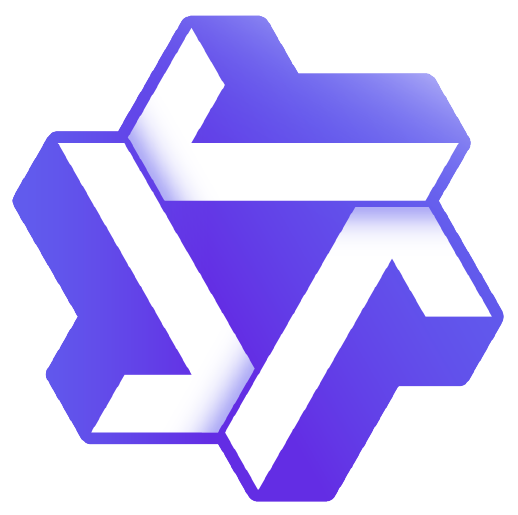}, Alibaba Group \\
  [0.8em]
  \texttt{\{shinji, zhaiyunpeng.zyp, eric.czq, chengchen, zouanni.zan, taoshuchang.tsc, jingmu.lzy, bolin.ding\}@alibaba-inc.com}
}

\begin{document}
\setlength{\floatsep}{6pt plus 2pt minus 2pt}
\setlength{\textfloatsep}{6pt plus 2pt minus 2pt}
\setlength{\intextsep}{6pt plus 2pt minus 2pt}

\maketitle

\begingroup
  \renewcommand\thefootnote{}%
  \footnotetext{%
    \begin{tabular}{@{}l@{\hspace{0.4em}}l@{}}
      $^{*}$ & Equal contribution.\\
      $^{\dagger}$ & Corresponding authors.
    \end{tabular}%
  }%
\endgroup


\begin{abstract}
Large language model (LLM)–based agents are increasingly deployed in complex, tool-augmented environments. While reinforcement learning (RL) provides a principled mechanism for such agents to improve through interaction, its effectiveness critically depends on the availability of structured training tasks. In many realistic settings, however, no such tasks exist—a challenge we term task scarcity, which has become a key bottleneck for scaling agentic RL. Existing approaches typically assume predefined task collections, an assumption that fails in novel environments where tool semantics and affordances are initially unknown. To address this limitation, we formalize the problem of Task Generation for Agentic RL, where an agent must learn within a given environment that lacks predefined tasks. We propose \textbf{CuES}, a \textbf{Cu}riosity-driven and \textbf{E}nvironment-grounded \textbf{S}ynthesis framework that autonomously generates diverse, executable, and meaningful tasks directly from the environment’s structure and affordances, without relying on handcrafted seeds or external corpora. CuES drives exploration through intrinsic curiosity, abstracts interaction patterns into reusable task schemas, and refines them through lightweight top-down guidance and memory-based quality control. Across three representative environments—AppWorld, BFCL, and WebShop—CuES produces task distributions that match or surpass manually curated datasets in both diversity and executability, yielding substantial downstream policy improvements. These results demonstrate that curiosity-driven, environment-grounded task generation provides a scalable foundation for agents that not only learn how to act, but also learn what to learn. The code is available at \url{https://github.com/modelscope/AgentEvolver/tree/main/research/CuES}.
\end{abstract}


\section{Introduction}

Large language model (LLM)–based agents are increasingly deployed in complex, tool-augmented environments such as graphical user interfaces and open-ended web platforms \citep{shang2025rstar2, li2025system}. Reinforcement learning (RL) provides a principled mechanism for such agents to improve their policies through interaction and feedback \citep{mai2025agent, lin2025comprehensive}. However, in many realistic environments, \textbf{no explicit training tasks are available}. The agent can interact with the environment, but it lacks a structured set of tasks on which it can sample experience and perform RL optimization \citep{cheng2024seeclick,zhang2024agentohana}. As a result, this \textit{task scarcity} has become a major bottleneck for scaling RL-based agent learning.

Existing approaches to agentic RL typically assume that a collection of training tasks is already provided \citep{gao2024agentscope, yu2025demystifying, chinta2020agentic}. These assumptions hold in controlled benchmarks but fail in novel environments, where tool semantics and affordances are initially unknown. In such cases, constructing a sufficiently diverse and executable set of tasks by hand is both costly and brittle, and the absence of such tasks leaves the agent without a foundation for effective learning \citep{zhang2025agentrl, zeng2025glm}. Consequently, the lack of training tasks—not the lack of algorithms—often limits progress in agentic RL\citep{plaat2025agentic, singh2025agentic, sapkota2025ai}.

To address this fundamental limitation \citep{shi2025taskcraft,team2025kimi}, we focus on the overlooked yet essential setting of \textbf{training an agent within a given environment when no tasks are predefined}. Specifically, we aim to answer the following question: \textit{Given an interactive environment but no predefined training tasks, how can an agent autonomously generate diverse, solvable, and useful tasks that enable effective policy learning?} To this end, we first \textbf{formulate and define the task generation problem} for such environment-conditioned learning. We conceptualize \textit{task generation} as the process of constructing a meaningful and solvable set of tasks directly from the environment’s structure and affordances, thereby bridging the gap between the environment and the agent’s learning process. This formulation establishes task generation as a first-class research problem in the development of self-improving LLM-based agents.

Building on this formulation, we propose \textbf{CuES}, a \textbf{Cu}riosity-driven and \textbf{E}nvironment-grounded \textbf{S}ynthesis framework for autonomous task generation. CuES operates without handcrafted seed goals or external corpora. It drives the agent to explore the environment under intrinsic curiosity, abstracts discovered interaction patterns into reusable task schemas, and refines them using lightweight top-down guidance and memory-based quality control. Through this process, CuES generates executable and diverse training tasks that are naturally aligned with the agent’s learning needs, thereby enabling continual and scalable improvement through RL.

We evaluate CuES across three representative environments—AppWorld, BFCL, and WebShop—and find that the generated tasks match or surpass manually curated datasets in both diversity and executability. Moreover, when integrated into RL training, these tasks lead to substantial downstream policy improvement, demonstrating that curiosity-driven, environment-grounded task generation can effectively replace costly human task design.

Our main contributions are threefold:
\begin{itemize}
    \item \textbf{Problem Formulation.} We systematically analyze and formalize the problem of \textbf{Task Generation for Agentic RL}, where an agent must learn within a given environment that lacks predefined tasks. 
    \item \textbf{Method.} We propose \textbf{CuES}, a curiosity-driven and environment-grounded framework via \textbf{Bottom-Up Exploration} and \textbf{Top-Down Guidance}, which autonomously generates executable and diverse training tasks without relying on handcrafted seeds or external data.
    \item \textbf{Empirical validation.} Across multiple environments, CuES produces high-quality task distributions and strong downstream training performance, \textbf{confirming the effectiveness} of autonomous task generation for agentic learning.
\end{itemize}

Together, these contributions advance a systematic understanding and practical solution for learning in task-scarce environments—toward agents that not only learn \textit{how} to act, but also learn \textit{what} to learn.

\begin{figure}[t]
    \centering
    \includegraphics[width=0.8\linewidth]{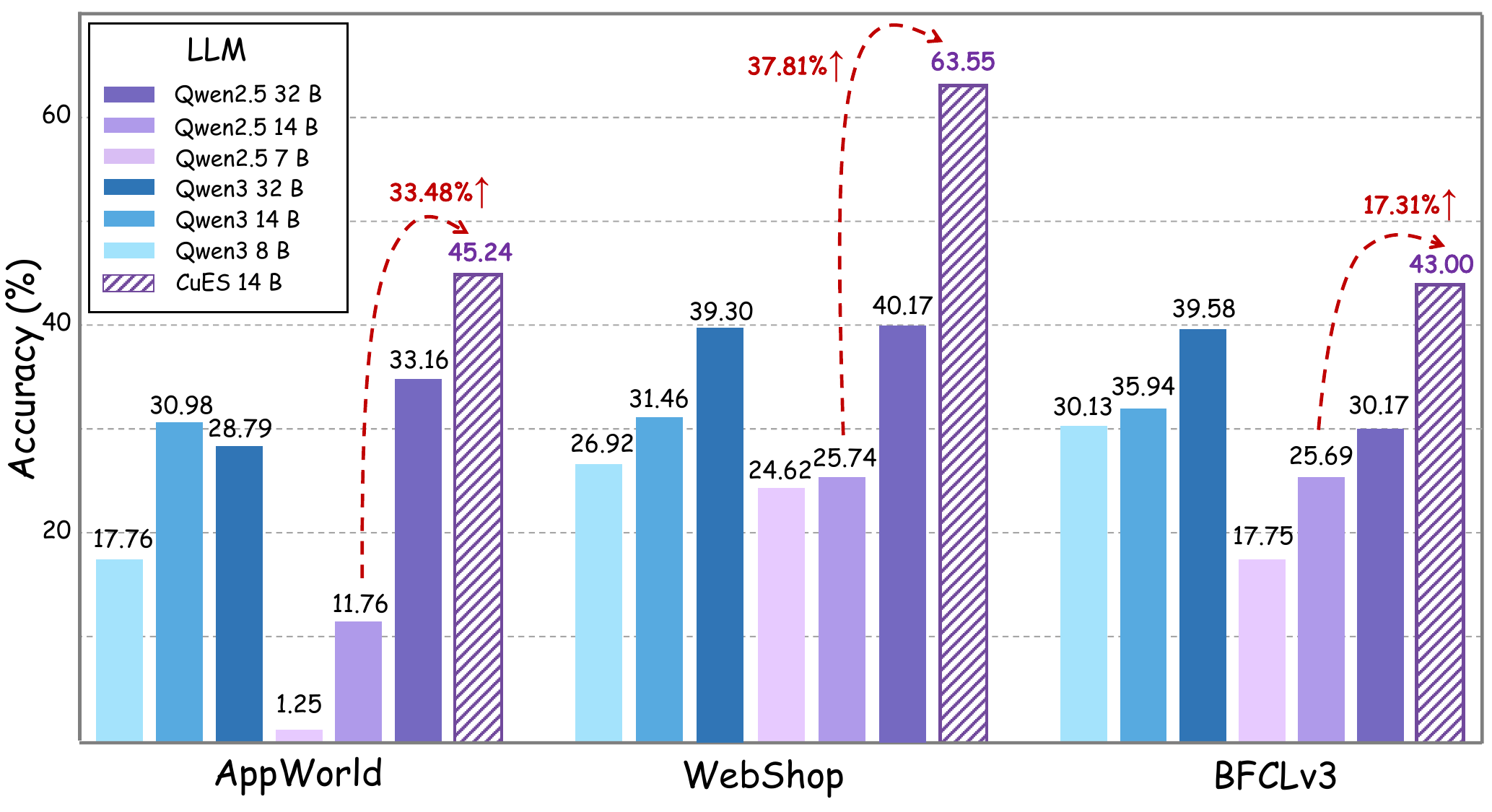}
    \caption{
    While performance on benchmarks(\textbf{AppWorld}, \textbf{WebShop}, and \textbf{BFCLv3}) continues to improve with larger LLMs, Qwen2.5 14B under the proposed \textbf{CuES} achieves a substantially higher accuracy across all benchmarks.
    }
    \label{fig:motivation}
\end{figure}

\section{Related Work}
To address this data scarcity, existing work generally follows two complementary lines. \textbf{Top-down approaches} start from human-written seed goals or LLM proposals and expand them into task instructions. This strategy benefits from clear goal specification and controllable semantics, but often \emph{decouples task generation from environment dynamics}, producing instructions that look plausible in text yet fail during execution\citep{li2024uieffects,lu2024weblinx,lai2024autowebglm}. In contrast, \textbf{bottom-up approaches} explore the environment first and then abstract tasks from discovered trajectories, ensuring that proposed problems are feasible in situ. While this improves executability and behavioral grounding, it typically suffers from weak goal alignment, inefficient exploration, and domain-specific heuristics that limit transferability. Moreover, the current exploration of bottom-up approaches only targets a few narrow areas such as web search, and the implementation is also highly coupled to specific tasks, which makes it difficult to be applied in practice\citep{sun2024osgenesis}. 

We organize related work into two threads: (i) top-down imitation-based synthesis that mimics existing data, (ii) bottom-up exploration that discovers tasks from interaction. CuES sits at the intersection: it is goal-free and bottom-up in how tasks emerge from interaction, while injecting lightweight top-down guidance via Environment Memory and Concept Pools, and enforcing executability with explicit judging.

\subsection{Top-down imitation-based synthesis: mimicking existing data}

Top-down pipelines typically rely on human-authored tasks or expand a small seed goal set using LLMs\citep{wu2025webdancer,li2025websailor, tao2025webshaper}. Recent web and GUI agents illustrate both the promise and the pitfalls of this approach. Systems such as AutoWebGLM \citep{lai2024autowebglm} and WebLINX \citep{lu2024weblinx} operate with explicit high-level goals or dialogue instructions and often imitate patterns present in existing benchmarks. While this improves goal clarity, it tends to decouple proposals from environment dynamics, leading to low executability when intermediate preconditions are missing or tool affordances are misidentified. Empirical analyses show that scale alone does not resolve these issues: UI control studies find that simply increasing dataset size does not guarantee coherent, solvable trajectories \citep{li2024uieffects}, and self-improvement strategies without grounded verification may propagate subtle errors across steps \citep{patel2024selfimprove}.

The core limitations of imitation-based synthesis are twofold. First, dependence on predefined high-level tasks restricts scalability and curtails diversity, since the proposal space is ultimately bounded by seed goal patterns \citep{lai2024autowebglm}. Second, quality is hard to ensure: early-step mistakes or mismatched goals can corrupt entire trajectories, yielding incomplete or incoherent data even when text instructions look plausible \citep{li2024uieffects,patel2024selfimprove}. These issues form a bottleneck for advancing agents from scripted automation to robust autonomy in real environments.

CuES differs by removing reliance on manual seed goals and decoupling quality from imitation. Tasks emerge bottom up from witnessed interactions, then pass through explicit execution and judgment before any rewriting. Lightweight top-down signals—requirement confirmation and concept pools—serve to shape coverage and reduce wasted exploration, not to prescribe solutions. The result is a synthesis pipeline that attains the clarity often associated with top-down methods while maintaining the executability and groundedness of bottom-up discovery.

\subsection{Bottom-up exploration: discovering tasks from interaction}

Recent research has attempted to synthesize data directly from environmental interactions under specific benchmarks, overturning the traditional process of "specifying the task first, then collecting the trajectory". OS-Genesis \citep{sun2024osgenesis} exemplifies this reversal for GUI agents: an agent first explores the desktop and records raw traces, then retrospectively derives high-level tasks from the discovered behaviors, filtering trajectories to ensure quality and solvability. By grounding tasks in what the environment can actually support, this paradigm improves executability and increases variety relative to task-first scripting. These systems highlight a central advantage of bottom-up synthesis: proposals are feasible by construction because they originate from witnessed interactions.

Despite these gains, bottom-up pipelines face key challenges. Without sufficient guidance, exploration can drift, producing many trajectories that lack clear purpose or fail midway, which lowers overall yield and inflates redundancy \citep{murty2024bagel}. Moreover, many implementations are tailored to specific domains (e.g., GUI/web navigation), raising concerns about cross-domain generalization and transferability of the synthesized corpora \citep{sun2024osgenesis}. In practice, the absence of simple mechanisms to align discovery with desired coverage or stage-specific training needs can lead to substantial ineffective exploration and narrow gains outside the source domain.

CuES preserves the executability benefits of bottom-up discovery while addressing these limitations. It remains goal-free and interaction-driven but introduces requirement confirmation and concept pools to softly aim exploration toward salient regions of the environment without prescribing exact tasks. An environment-indexed memory stores compact state–action sketches, enabling the system to revisit informative frontiers and avoid redundant loops. Crucially, only validated successes proceed to rewriting, which broadens surface diversity while keeping the underlying behaviors executable.

\begin{figure}[t]
\centering
\includegraphics[width=6.2in]{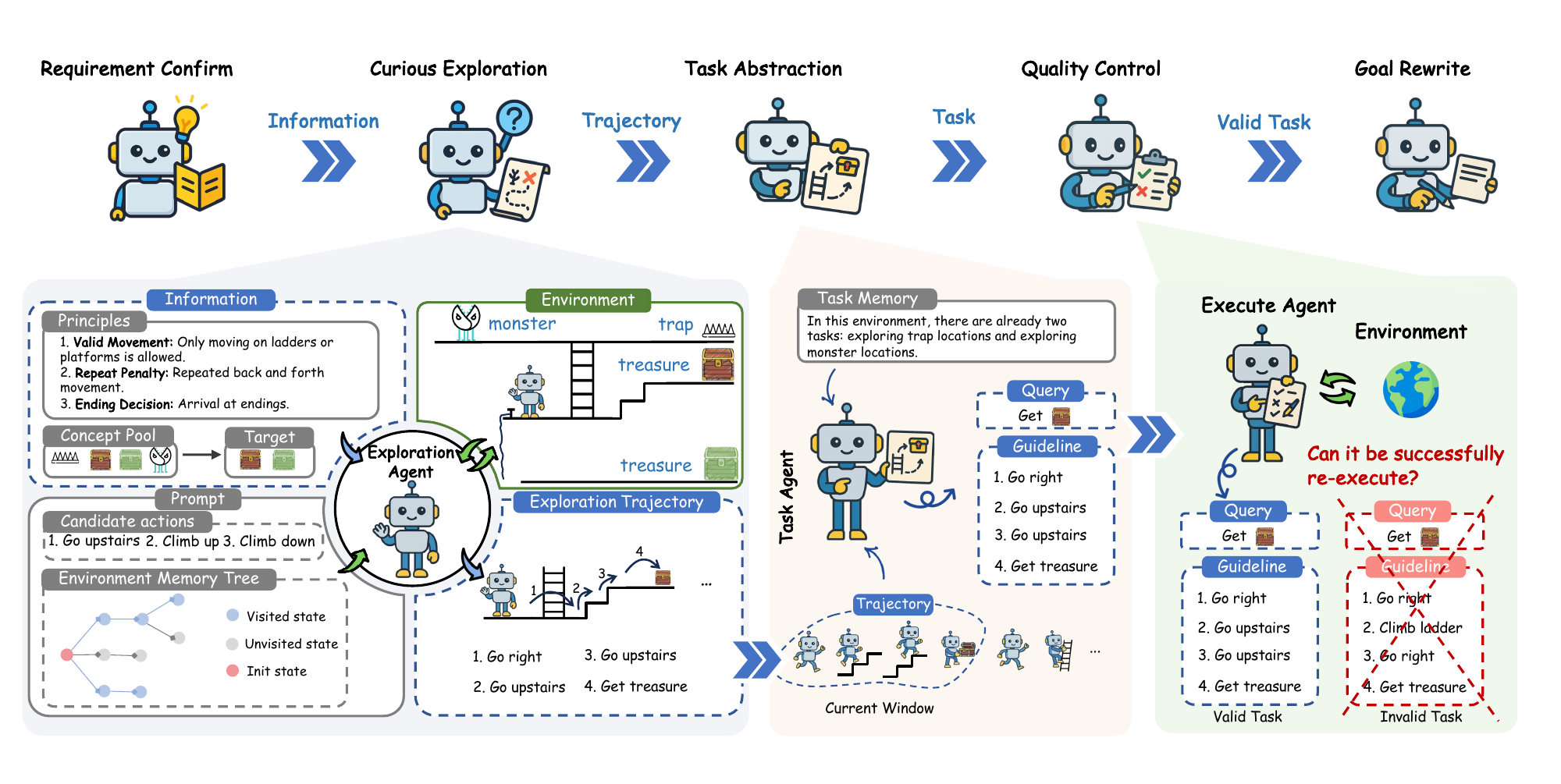}
\caption{CuES pipeline. (a) Requirement Confirm constructs the concept pool \(\tilde{\mathcal{C}}\) and principle \(\mathit{P}\) by extracting concepts from the environment description $T_{des}$ and seed goals $\mathcal{G}_{seed}$ and filtering them with the user need $\mathcal{U}$. (b) Curious Exploration executes candidate actions conditioned on $(\mathit{P},\mathcal{G}_{seed})$, consults the environment memory tree to prioritize unseen actions, and emits triples $(s,a,o)$ (eq.\ref{eq:triple}) as exploration trajectories. (c) Task Abstraction groups consecutive triples within a batch into executable goals with guidelines. (d) Quality Control re-executes each goal. (e) Goal Rewrite progressively exposes guideline hints in the goal text to lower difficulty.}
\label{fig:AgentFlowPipeline}
\end{figure}

\section{Formulation}

\label{sec:formu}

We consider the problem of enabling an LLM-based agent to learn in complex, interactive environments without access to predefined task distributions.
To differentiate our setting from conventional RL, we introduce an \emph{interaction sandbox} that specifies the observable state space, the executable action space, and transition dynamics, while deliberately omitting any predefined reward function or task objective.

\paragraph{Interaction sandbox.}
Formally, the sandboxed environment is
\begin{equation}
\label{eq:env}
\mathcal{E} = (\mathcal{S}, \mathcal{A}, \mathcal{P}),
\end{equation}
where $\mathcal{S}$ denotes the set of observable states, $\mathcal{A}$ the set of executable actions, and $\mathcal{P}(s' \mid s, a)$ the transition probability distribution.
In contrast to a standard Markov Decision Process $\mathcal{E}_{\mathrm{mdp}} = (\mathcal{S}, \mathcal{A}, \mathcal{P}, r, \gamma)$, the sandbox $\mathcal{E}$ \emph{omits} the reward function, characterizing an open-ended and non-rewarded environment in which the agent must autonomously formulate learning signals and objectives.

\paragraph{Target objective.}
Let $g\!\in\!\mathcal{G}\!\subseteq\!\mathcal{S}$ denote a desired terminal state or outcome, and $p_{\mathrm{target}}(g)$ the (unknown) target goal distribution.
For each $g$, $R_g(s,a)$ is a goal-specific reward (unknown during training).
Starting from $s_0\!\sim\!p_0$, a goal-conditioned policy $\pi_\theta(a\mid s,g)$ aims to maximize
\begin{equation}
\label{eq:target_objective}
J_{\mathrm{target}}(\theta)
=
\mathbb{E}_{g\sim p_{\mathrm{target}},\,s_0\sim p_0}\!\left[
V^{\pi_\theta}(s_0,g)
\right],
\quad
V^{\pi_\theta}(s_0,g)
=
\mathbb{E}\!\left[\sum_{t=0}^{\infty}\gamma^t\,R_g(s_t,a_t)\ \middle|\ s_0,g,\pi_\theta\right].
\end{equation}
Because $p_{\mathrm{target}}$ and $R_g$ are unavailable during training, directly optimizing $J_{\mathrm{target}}$ is infeasible.

\paragraph{Proxy goals.}
We induce a \emph{trainable} goal distribution from the environment by defining the task mapping
\begin{equation}
\label{eq:ftask}
F_{\text{task}}:\ \mathcal{E} \to \Delta(\mathcal{G}), \quad \text{where} \quad p_{\text{train}} = F_{\text{task}}(\mathcal{E}),
\end{equation}

We then optimize the proxy objective on $p_{\text{train}}$:
\begin{equation}
\label{eq:train_objective}
J_{\text{train}}(\theta) = \mathbb{E}_{g \sim F_{\text{task}}(\mathcal{E})} \left[ V^{\pi_\theta}(s_0, g) \right],
\end{equation}
which serves as the \emph{Proxy Objective for Environment-Conditioned Agentic Learning}.

\paragraph{Task mapping Design.}
$F_{\text{task}}$ fixes $p_{\text{train}}$ and thus the learning signal; to make $J_{\text{train}}$ informative for $J_{\mathrm{target}}$, we shape $F_{\text{task}}$ along three requirements:
\begin{itemize}
\item \textbf{Executability} ensures proposed goals are admissible and terminate in recognizable outcomes; this suppresses noisy supervision and stabilizes updates. 
\item \textbf{Diversity} spreads probability mass across distinct entities, actions, and constraints; this enlarges support and mitigates collapse to easy templates. 
\item \textbf{Relevance} keeps synthesized goals close in semantics and difficulty to the intended evaluation targets; this prevents drift toward off-topic but easy objectives. 
\end{itemize}

Taken together, these three requirements turn $F_{\text{task}}$ into a principled mapper that yields proxy goals which are valid (executability), broad (diversity), and faithful to evaluation (relevance). This improves stability, coverage, and transfer of the learning signal.

\section{Method}

\label{sec:method}

Under the formulation, we clearly defined the Proxy Objective for Environment-Conditioned Agentic Learning, and we found that the most important core task mapping Eq.~\eqref{eq:ftask} limited Agentic Learning. Therefore, we need to implement the mapping $F_{\mathrm{task}}$ to close the loop between data and policy.

We operationalize $F_{\mathrm{task}}$ via \textbf{CuES}, a \textbf{Cu}riosity-driven and \textbf{E}nvironment-grounded framework for agentic data \textbf{S}ynthesis mapping in goal-free or limited goal environments:
\begin{equation}
\label{eq:fcues}
F_{\text{task}}:\mathcal{E}, (T_{des},T_{req},\mathcal{G}_{seed})\ \xrightarrow[]{\textbf{CuES}}\ \mathcal{G}_{\mathrm{synthesis}}, 
\end{equation}

where $\mathcal{E}$ is an \textbf{E}xecutable environment including a finite tool set in CuES. $T_{des}$ is a structured \textbf{T}ext of environment \textbf{des}cription, $T_{req}$ is an optional \textbf{T}ext of user \textbf{req}uirement, and $\mathcal{G}_{seed}$ is an optional \textbf{seed}-goal set. 

Based on the analysis and derivation in the formulation, implementing $F_{\mathrm{task}}$ also requires the design rules of Executability, Diversity, and Relevance mentioned in Sec.~\ref{sec:formu}. As illustrated in Fig.~\ref{fig:AgentFlowPipeline}, CuES generates tasks with \textbf{Executability}, \textbf{Diversity}, and \textbf{Relevance} through a five-stage design process. 

\subsection{Requirement Confirmation}
\label{sec:req}

To secure \textbf{relevance} at the outset, before the agent begins exploring, it must build a structural understanding of the environment, including the entities, the actions, and the admissible interactions under given constraints. Without this grounding, curiosity drifts into random or redundant behaviors that are executable but off target, weakening relevance. Therefore, CuES introduce Requirement Confirmation stage.

As shown in Fig.~\ref{fig:AgentFlowPipeline}(a), the requirement confirmation stage takes three inputs: an environment description $T_{des}$, an optional filter agent $\mathcal{U}$ based on user requirement $T_{req}$, and an optional seed goal set $\mathcal{G}_{seed}$. This also demonstrates a major advantage of CuES. Seed goal set is required for almost all synthesis methods, but in CuES, seed goal set and other requirements, including user requirement, are not mandatory and are not provided by default. They are only provided when users are only concerned with the use of a certain type of tool and want to provide strong constraints to improve exploration efficiency. Requirement confirmation stage produces two outputs: a \textbf{concept pool} $\tilde{\mathcal{C}}$, which grounds subsequent exploration, and a set of \textbf{actionable principles} $\mathit{P}$, which is extracted by principle agent through environment description and concept pool and specify output schema and highlight priority actions. Both outputs are passed forward to guide the next stage.

To initialize exploration, we construct a preliminary concept pool by combining concepts extracted from the environment and from the seed set:
\begin{equation} 
\tilde{\mathcal{C}} \;=\; \mathcal{U}\big(\Phi(T_{des}) \;\cup\; \Phi(\mathcal{G}_{seed})\big).
\end{equation}

Here, $\Phi(T_{des})$ identifies entities and affordances from $T_{des}$ (e.g., categories, tools, admissible actions), while $\Phi(\mathcal{G}_{seed})$ extracts noun phrases and action predicates when $\mathcal{G}_{seed}$ is provided. When filter agent $U$ is specified, it filters the pool to obtain the final concept set; otherwise, the preliminary pool remains unchanged. Detailed examples refer to Fig.\ref{fig:AgentFlowPipeline}(a).

Principles $\mathit{P}$ are carried forward into the next stage’s prompt to specify the desired output format and to highlight the main action families to explore, while $\tilde{\mathcal{C}}$ anchors exploration in the entities and actions admissible in the environment.

\subsection{Curious Exploration}
\label{sec:explore}

Before meaningful task patterns can emerge, the agent must actively explore to reveal new state–action paths that are truly solvable. Random or unguided search quickly becomes inefficient, revisiting similar states and missing unseen affordances, which lowers \textbf{diversity}. Thus, Curious Exploration stage is necessary.

As shown in Fig.\ref{fig:AgentFlowPipeline}(b), given the top-down gate formed by the principles $\mathit{P}$ and the concept pool $\tilde{\mathcal{C}}$, Curious Exploration stage conducts bottom-up interaction with an Explorer Agent that, at each state $s_t$, receives only the currently executable candidate actions $\mathcal{A}(s_t)$, together with $\mathit{P}$ and a small, randomly sampled subset of concepts $\tilde{\mathcal{C}}_t\subset\tilde{\mathcal{C}}$. 

To avoid redundant exploration and preserve local know-how, we maintain an environment-indexed memory that reuses what was tried before. The key idea is simple: distinguish environments by an identifier $\mathrm{env\_id}$, store the trajectories collected under each identifier together with concise summaries, and, upon re-entering the same identifier, retrieve these summaries to guide action selection.

Let $e$ be the environment identifier $env\_id$ extracted from the current environment state $s$. We define the memory tree as
\begin{equation}
\label{eq:memtree}
\mathbb{M}\;=\;\{(e,\mathsf{B}_e)\}, 
\qquad 
\mathsf{B}_e\;=\;\{\ \mathrm{Summ}(\tau)\ :\ \tau\in\mathcal{T}_e\ \},
\end{equation}
where $\mathcal{T}_e$ is the multiset of trajectories collected when $env\_id=e$, and $\mathrm{Summ}(\cdot)$ compresses a trajectory $\tau=(s_0,a_0,o_0,\ldots,s_T)$ into a lightweight record (e.g., executed actions, success/failure flags, brief outcome snippets, and counts).

The Explorer queries an environment memory tree $\mathbb{M}$ using the current environment identifier $s_t$ retrieving the set of actions $\mathbb{M}(s_t)$ previously executed at states similar to $s_t$.

The Explorer then decides $a_t\in\mathcal{A}(s_t)$, with a preference for actions not yet attempted at the local memory node (i.e., $a\notin\mathbb{M}(s_t)$, while taking into account the priorities expressed by $\mathit{P}$ and the active concepts $\tilde{\mathcal{C}}_t$.

After executing $a_t$, the environment returns an observation $o_t$. The memory tree is updated at node $\phi(s_t)$ to record the attempt and its outcome via a compact sketch:
\begin{equation}
\mathbb{M}(s_t) \leftarrow \mathbb{M}(s_t) \cup \{a_t\}
\end{equation}

The Explorer emits a triple per step,
\begin{equation}
z_t=(s_t,\;a_t,\;o_t),
\end{equation}
\label{eq:triple}

and the exploration trajectory is the sequence $\{z_t\}_{t=0}^{T}$. These concept-aware, memory-informed trajectories are forwarded to the next stage task abstraction in the pipeline, where low-level interactions are lifted into reusable task specifications while preserving the executability demonstrated during exploration.

\subsection{Task Abstraction}
\label{sec:abstract}

After exploration produces a collection of raw interaction trajectories, the next challenge is to transform these low-level stepwise records into meaningful, reusable training tasks, which ensures the \textbf{diversity} and \textbf{executability} of the task. Refer to Fig.\ref{fig:AgentFlowPipeline}(c), Task Abstraction stage takes a \emph{consecutive} mini-batch of exploration triples and lifts one or more multi-step tasks from it. Let a batch be
\begin{equation}
\Bigl\{\, 
\{(s_t, a_t, o_t)\}_{t=i}^{i+B-1}
\;\Big|\;
i \in \{1,\, 1{+}B,\, 1{+}2B,\, \dotsc\}
\,\Bigr\},
\end{equation}
where $B=\texttt{batch\_size}$ denotes the sliding-window length. Each triple consists of the step state $s_t$, the executed action $a_t$, and the resulting observation $o_t$. Rather than extracting a goal from every triple, we consider \emph{contiguous segments} inside the batch.

For each batch starting at $t_k$, we then enumerate all consecutive subsegments within it,
$
\bigl\{\, [i{:}j] \;\big|\;
t_k \le i < j \le t_k{+}B{-}1 \,\bigr\},
$
where each $[i{:}j]$ corresponds to a contiguous slice of triples. For a candidate segment $[i{:}j]$, Task Agent extracts a new goal valid action sequence $z_{i:j}$ which is also called \emph{guideline} and executable goal $g_{i:j}$ which is a natural-language rendering of $z_{i:j}$. For LLM judge, Directly judging whether a trajectory is correct and giving a reward close to $R_g$ is too difficult, guideline is used to assist in the determination. It will be used as $\hat{R}_g = R(g_{i:j}, z_{i:j})$ to estimate $R_g$.

Each candidate $(g_{i:j},z_{i:j})$ receives a confidence $\sigma_{i:j}\in[0,1]$ that reflects internal consistency of the action–effect chain, clarity of the goal, and alignment with current principles and concepts. 
We keep those $\mathcal{G}$ that pass the threshold.

To avoid redundancy, for each environment identifier we maintain a memory list of previously generated goals. When proposing a new goal, we form a context and supply during goal generation so that duplicates are explicitly flagged as already generated.

\begin{figure*}[t]
\centering

\begin{subfigure}{0.5\linewidth}
\centering
\includegraphics[width=\linewidth]{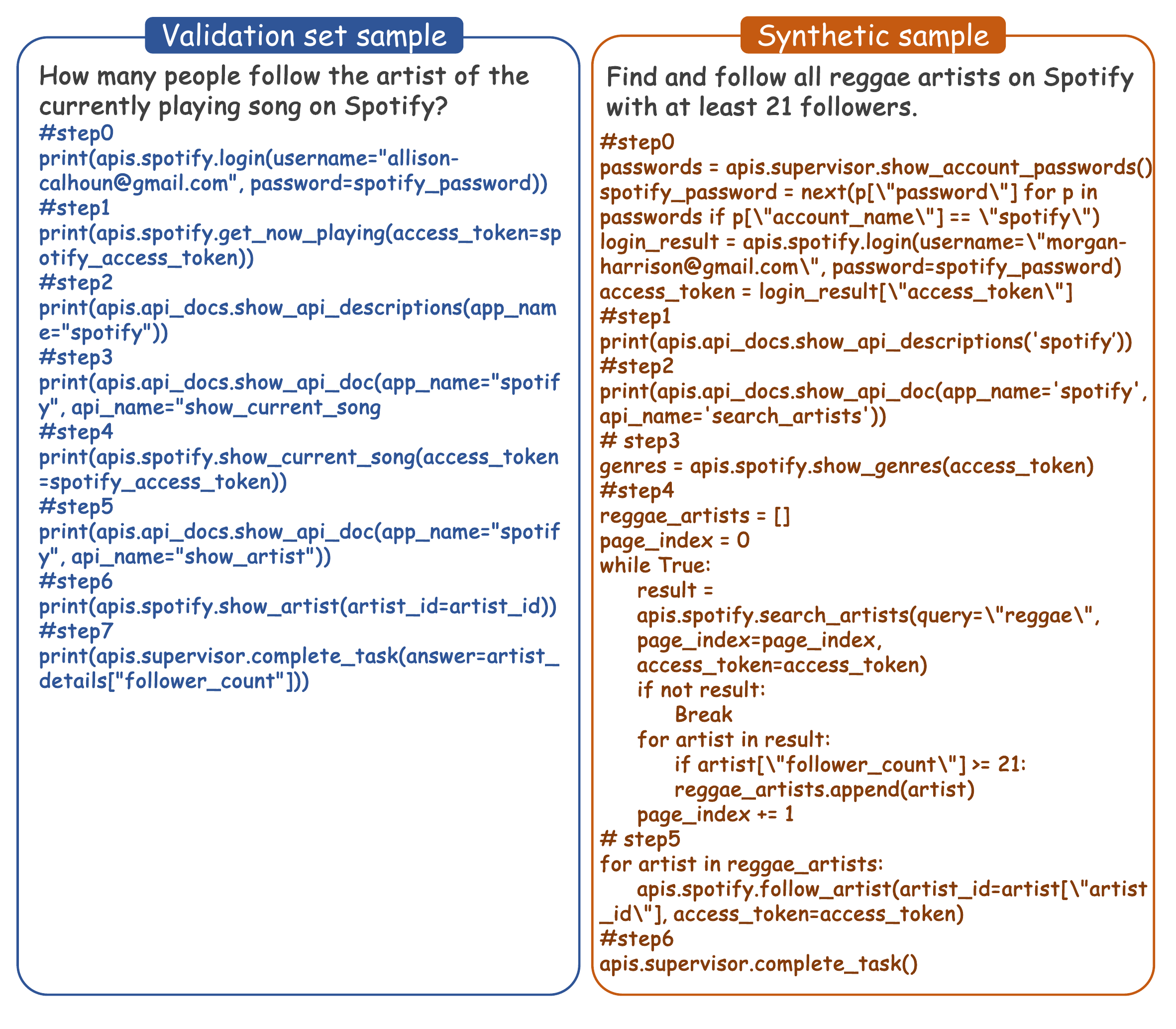}
\caption{AppWorld}
\end{subfigure}\hfill
\begin{subfigure}{0.5\linewidth}
\centering
\includegraphics[width=\linewidth]{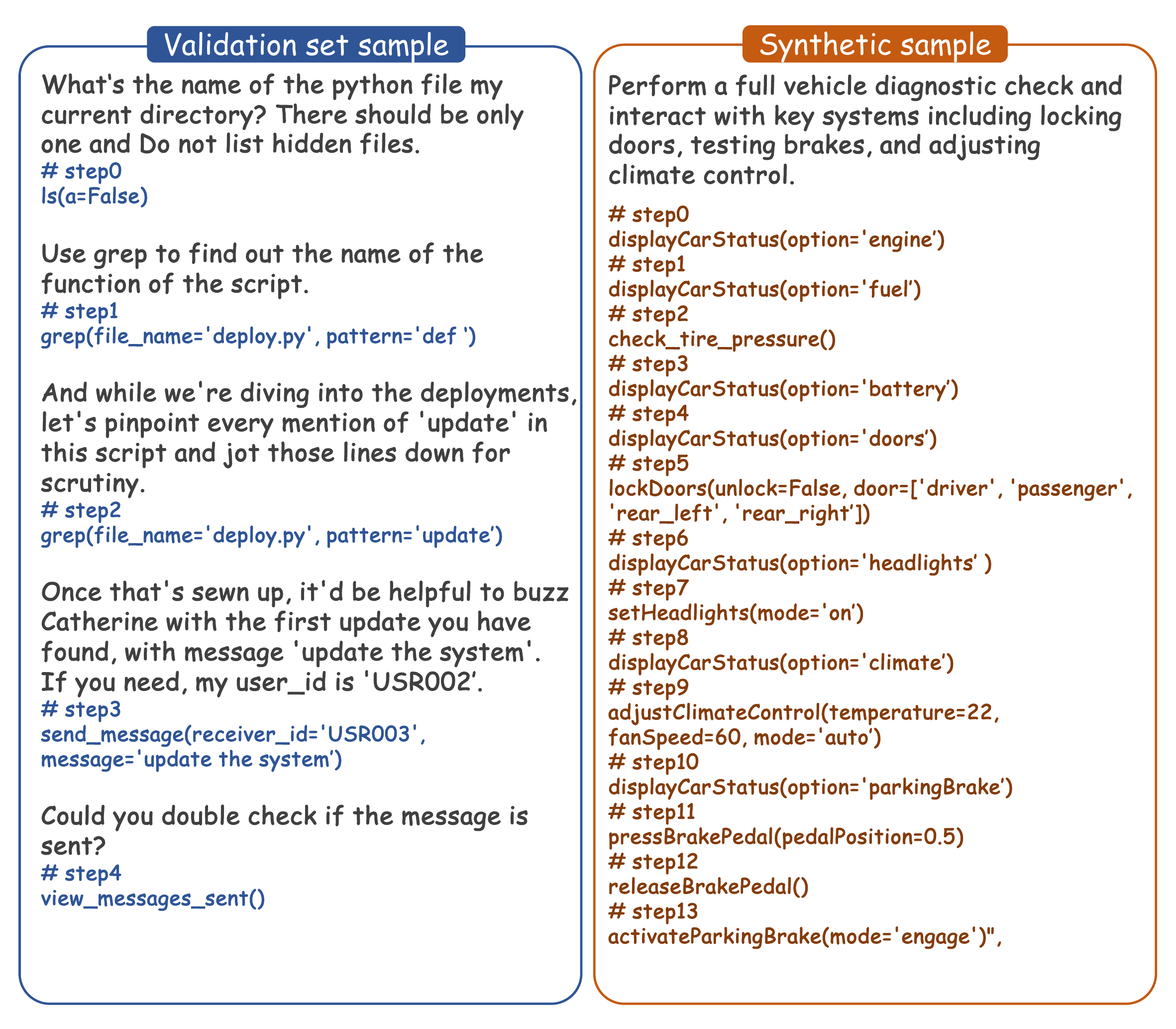}
\caption{BFCL v3 Multi-Turn Base}
\end{subfigure}

\caption{Original (left) vs CuES-synthesized (right) data for AppWorld and BFCL v3 Multi-Turn Base. }
\label{fig:samples}
\end{figure*}

\subsection{Quality Control}
\label{sec:qc}

Even after abstraction, not all synthesized tasks are guaranteed to be executable—some may contain invalid actions, incomplete goals, or inconsistencies between intent and behavior. Training on such noisy or unverified data would undermine the reliability of downstream learning. In Fig.\ref{fig:AgentFlowPipeline}(d), Quality Control stage verifies the \textbf{executability} of each synthesized task using two agents. Let the input be the candidate set from task abstraction $\{\, (g_k,\ z_k\}_{k=1}^{N} \}$, where $g_k$ is a natural-language goal, $z_k=[a_{i_k},\dots,a_{j_k}]$ is the associated action guideline extracted from a contiguous segment.

Execution Agent receives $(g_k,z_k)$ and attempts to carry out the task in environment $\mathcal{E}$, following the guideline actions in order while allowing minimal, environment-dependent deviations. The attempt yields an execution trace $\big((\tilde{s}_0,\tilde{a}_0,\tilde{o}_0),\ldots,(\tilde{s}_{T_k},\tilde{a}_{T_k},\tilde{o}_{T_k})\big)$ with terminal observation. The trace, together with local outcomes (success messages, view changes), is returned for judgment.

Judge Agent checks both goal satisfaction and path faithfulness and verifies that the observed actions respect the principles $\mathit{P}$ and guideline up to allowable, environment-specific insertions (for example, a short detour to grant permission). A candidate is \emph{accepted} if $reward=1.0$ and \emph{rejected} otherwise.

For accepted tasks, we log both the successful goal and its execution trajectory, so that subsequent stages and future explorations have access to validated problem statements and their witnessed solutions. Rejected candidates are filtered out and do not proceed further. Only the accepted set $\mathcal{G}_{\mathrm{original}}$ flows to goal rewrite) stage, ensuring that downstream diversification operates exclusively on tasks that have been explicitly executed to completion under their accompanying guidelines.
\vspace{2mm}
\begin{figure*}[htbp]
\centering
\begin{subfigure}{0.32\linewidth}
\centering
\includegraphics[width=\linewidth]{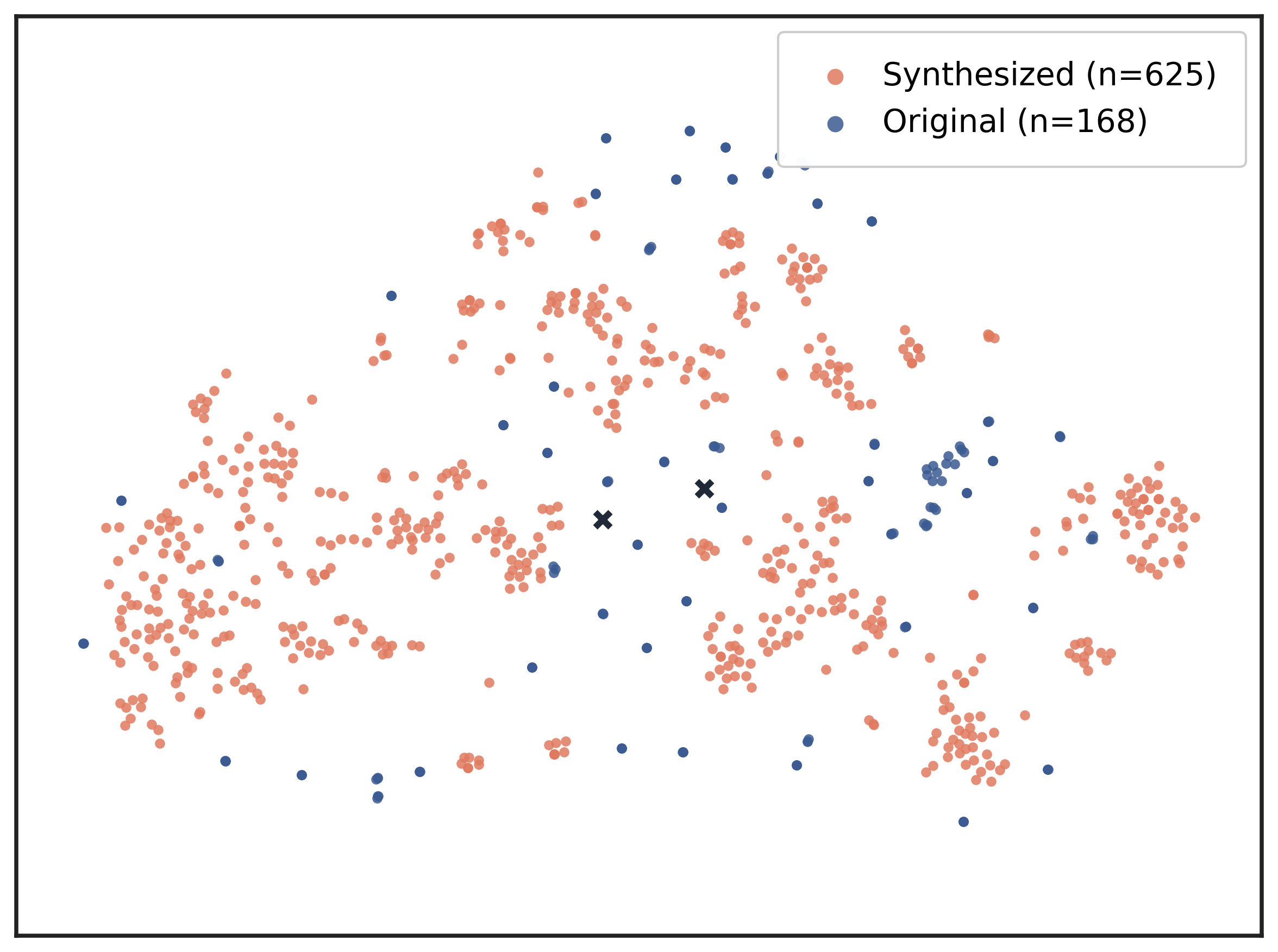}
\caption{\textsc{AppWorld}}
\end{subfigure}\hfill
\begin{subfigure}{0.32\linewidth}
\centering
\includegraphics[width=\linewidth]{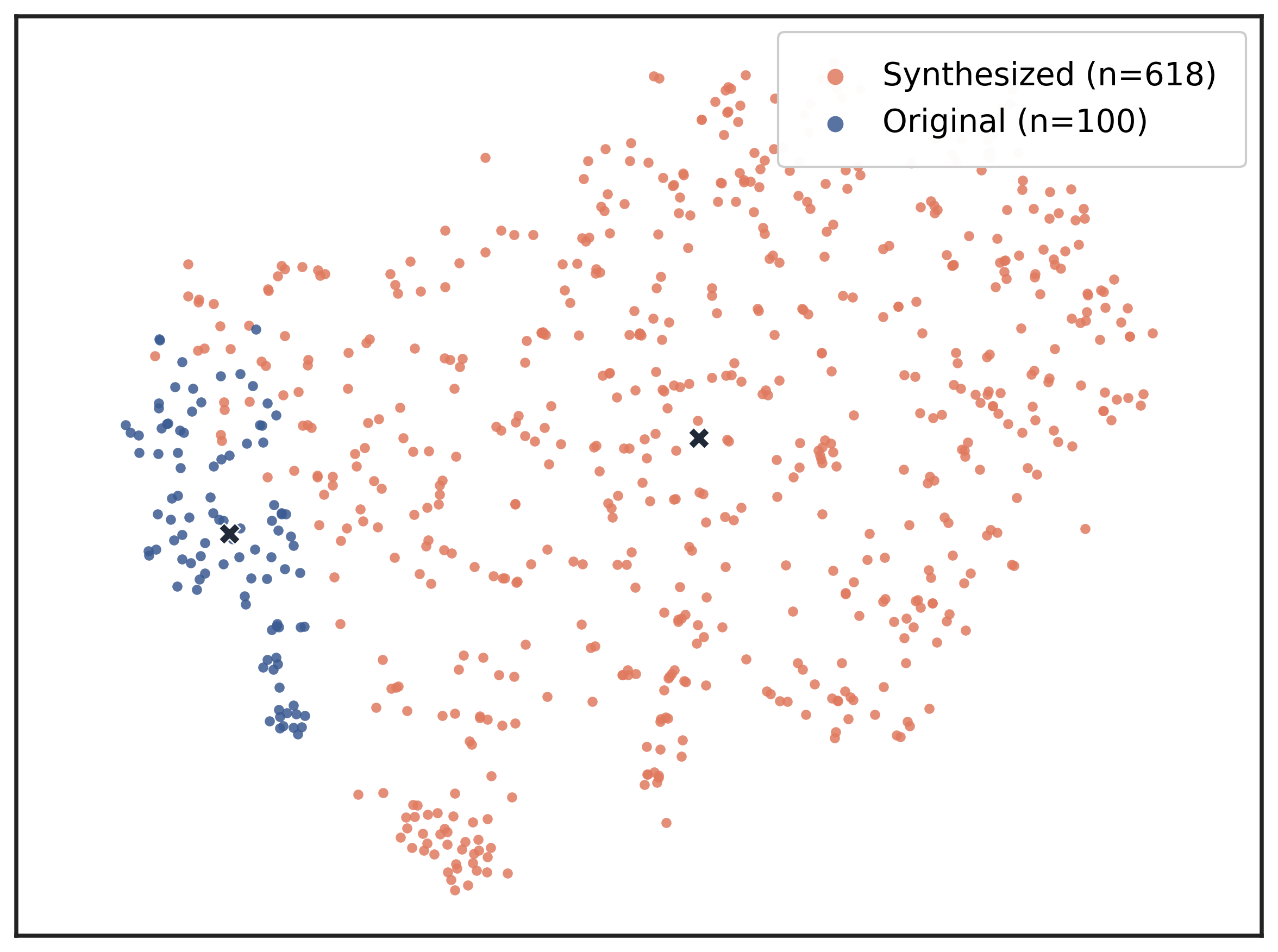}
\caption{\textsc{WebShop}}
\end{subfigure}\hfill
\begin{subfigure}{0.32\linewidth}
\centering
\includegraphics[width=\linewidth]{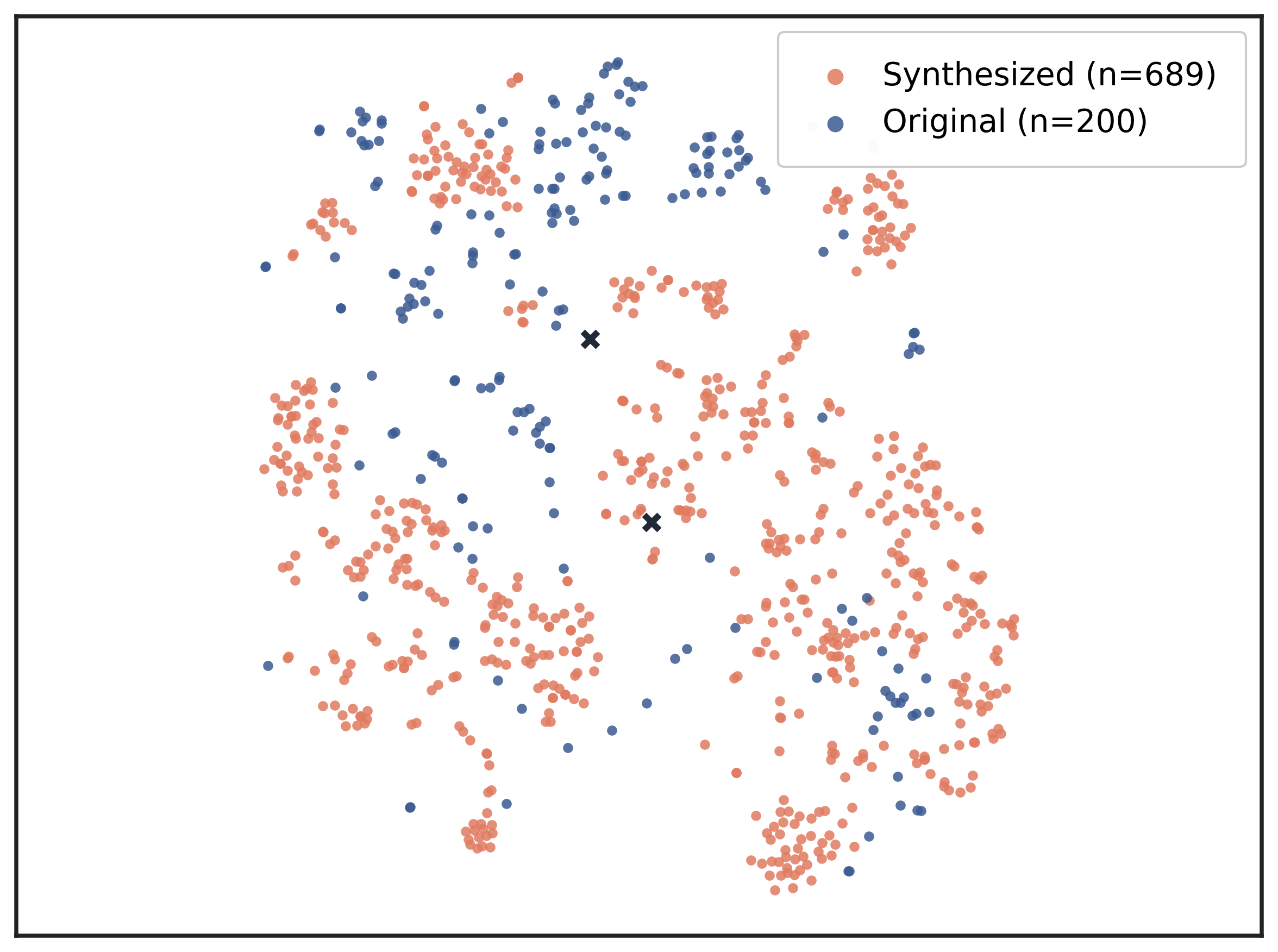}
\caption{\textsc{BFCL v3 Multi-Turn Base}}
\end{subfigure}

\caption{Distribution comparison per environment.}
\label{fig:dist_grid}
\end{figure*}

\subsection{Goal Rewrite}
\label{sec:rewrite}

Even after verification, the accepted tasks may vary greatly in complexity, making them unevenly useful for training. Some goals are overly concise or abstract, offering limited guidance for early-stage learning, while others are too detailed to challenge advanced policies. To build a balanced and curriculum-friendly dataset, we introduce the Goal Rewrite stage, which systematically adjusts task difficulty to ensure \textbf{diversity} and \textbf{relevance} by revealing the guideline portion into the goal.

Ref to Fig.~\ref{fig:AgentFlowPipeline}(e), the rewrite stage operates on the accepted set
\[
\mathcal{G}_{\mathrm{original}}=\big\{(g_k,\, z_k)\big\}_{k=1}^{N}.
\]
We use a rewrite-depth hyperparameter $L$ shared by all goals. For each $(g_k,z_k)$, let $\Gamma_k$ be a pool of rewrite hints distilled from $z_k$ (e.g., action verbs, intermediate landmarks/views, parameter exemplars, precondition CuES). We generate a chain of exactly $L$ steps:
\[
g_k^{(0)}=g_k,\quad g_k^{(1)},\ \dots,\ g_k^{(L)},
\]
via
\begin{equation}
\label{eq:addhints_fixed}
g_k^{(\ell+1)} \;=\; \mathrm{AddHints}\!\big(g_k^{(\ell)};\ \Delta \Gamma_k^{(\ell)}\big),
\qquad \Delta \Gamma_k^{(\ell)}\subseteq \Gamma_k,\ \ \Delta \Gamma_k^{(\ell)}\neq\varnothing,\ \ \ell=0,\dots,L-1.
\end{equation}
Every goal is rewritten for exactly $L$ steps under the same cap. Finally,
\begin{equation}
\label{eq:Gsynth_fixed}
\mathcal{G}_{\mathrm{synthesis}}
=\Big\{\,(g_k^{(\ell)},\, z_k)\ :\ k=1,\dots,N,\ \ell=0,\dots,L\,\Big\}.
\end{equation}

\begin{table*}[t]
\small
\renewcommand\arraystretch{0.8}
\centering
\caption{Model performance (\%) on three interactive environments using \emph{avg@8} and \emph{greedy} under the no-think and prompt configuration of Qwen2.5 and Qwen3}
\label{tab:scale_perf_compact}

\begingroup
\setlength{\tabcolsep}{1.4mm}
\begin{tabular}{l c cc cc cc cc}
\toprule
\multirow{2}{*}{\textbf{Model}} & \multirow{2}{*}{\textbf{Params}}
& \multicolumn{2}{c}{\textbf{AppWorld}}
& \multicolumn{2}{c}{\textbf{WebShop}}
& \multicolumn{2}{c}{\textbf{BFCLv3}}
& \multicolumn{2}{c}{\textbf{Avg.}} \\
\cmidrule(lr){3-4}\cmidrule(lr){5-6}\cmidrule(lr){7-8}\cmidrule(lr){9-10}
& & avg@8 & greedy & avg@8 & greedy & avg@8 & greedy & avg@8 & greedy \\
\midrule
Qwen2.5 & 3B   & 0.00\% & 0.23\% & 20.65\% & 23.94\% & 6.94\%  & 7.00\%  & 9.20\% & 10.39\% \\
Qwen2.5 & 7B   & 1.25\% & 1.87\% & 24.62\% & 22.07\% & 17.75\% & 20.00\% & 14.54\% & 14.65\% \\
Qwen2.5 & 14B  & 11.76\% & 14.29\% & 25.74\% & 23.74\% & 25.69\% & 31.50\% & 21.06\% & 23.18\% \\
Qwen2.5 & 32B  & \underline{33.16}\% & \underline{34.73}\% & \underline{40.17}\% & \underline{39.80}\% & 30.17\% & 30.50\%  & 34.50\% & 35.01\% \\
\midrule
Qwen3   & 4B   & 3.00\% & 3.25\% & 33.72\% & 33.17\% & 9.94\%  & 10.00\% & 15.55\% & 15.47\% \\
Qwen3   & 8B   & 17.76\% & 21.43\% & 26.92\% & 30.01\% & 30.13\% & 27.50\% & 17.81\% & 19.98\% \\
Qwen3   & 14B  & 30.98\% & 28.48\% & 31.46\% & 30.70\% & 35.94\% & 30.70\% & 32.79\% & 29.96\% \\
Qwen3   & 32B  & 28.79\% & 32.12\% & 39.30\% & 36.69\% & \underline{39.58}\% & \underline{41.00}\% & \underline{35.89}\% & \underline{36.60}\% \\
\rowcolor{gray!40}
\textbf{CUES (ours)} & 14B  & \textbf{45.54\%} & \textbf{45.24\%} & \textbf{63.55\%} & \textbf{64.10\%} & \textbf{43.00\%} & \textbf{44.15\%} & \textbf{50.70\%} & \textbf{51.16\%} \\
\bottomrule
\end{tabular}
\endgroup

\end{table*}

\section{Experiments}

\subsection{Experimental Setup}

\begin{wrapfigure}{r}{0.34\textwidth}
    \centering
    \includegraphics[width=\linewidth]{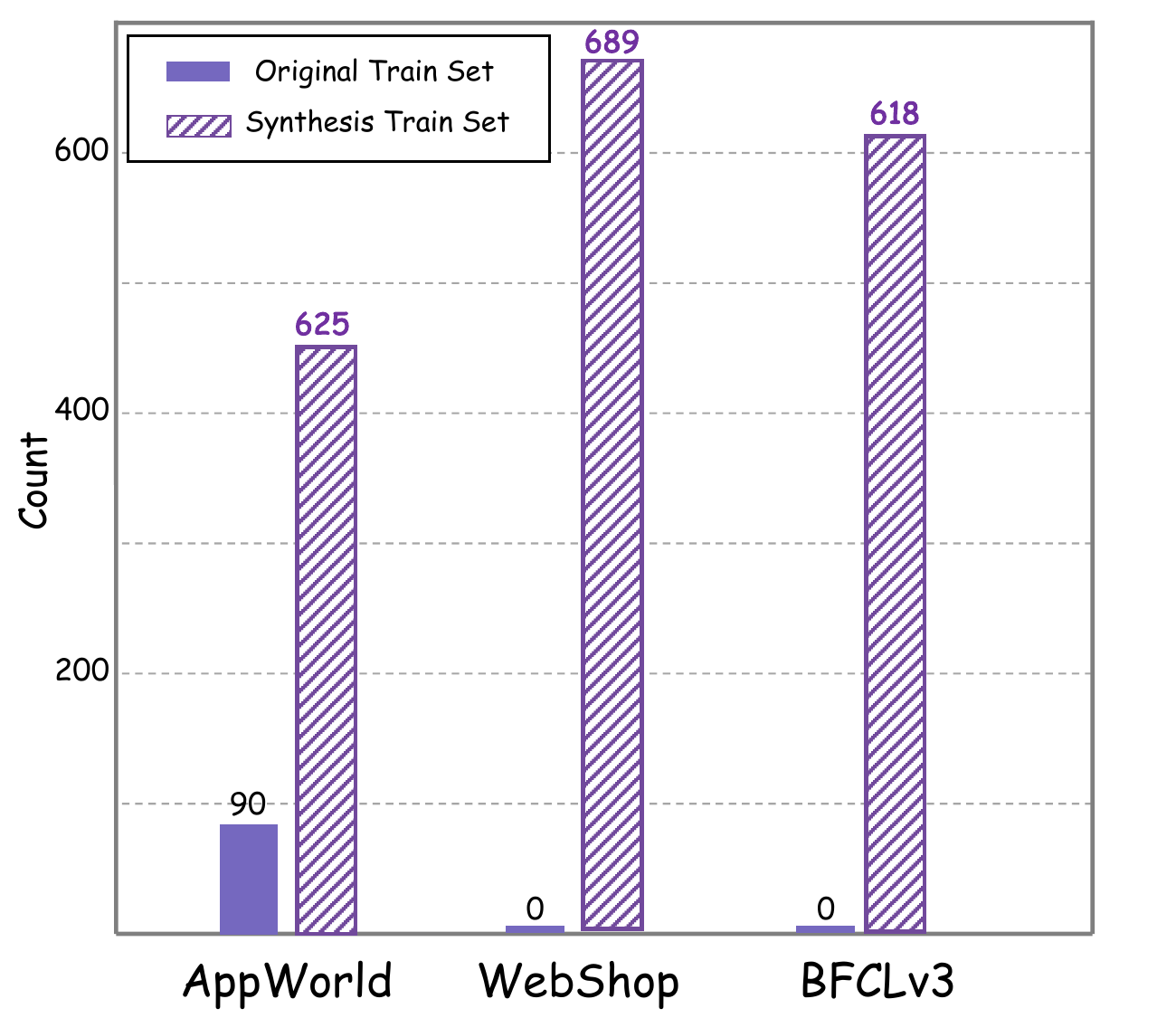}
    \caption{
    \textbf{Comparison of original and CuES-synthesized data.}
    }
    \label{fig:data_count}
    \vspace{-2mm}
\end{wrapfigure}

We evaluate CuES on three interactive environments: AppWorld, WebShop, and BFCL v3 Multi-Turn Base. For each environment, we report the size of the original data and the amount synthesized by CuES under a matched interaction budget, alongside a brief description of the core task. Unless otherwise noted, we use the same runtime across environments: \texttt{model\_name} = \texttt{qwen-plus-latest}, \texttt{temperature} = 0.7, \texttt{max\_tokens} = 20480; Curious Exploration Stage (\ref{sec:explore}) rollout count = 500 with up to 30 steps per rollout; Stage Task Abstraction (\ref{sec:abstract}) uses \texttt{batch\_size} = 30 and \texttt{min\_confidence} = 0.7; Stage Quality Control (\ref{sec:qc}) attempts up to 30 steps with \texttt{retry\_attempts} = 3; the environment service has a 30s timeout and 30-step cap.

For more information about the dataset, ref to Fig.\ref{fig:data_count}. For benchmarks that include validation sets, such as BFCL v3 and WebShop, we use the validation set directly. For AppWorld, we use test\_normal set as the test set. The reported results are equivalent to the TGC indicator on the leaderboard. In Fig.\ref{fig:samples}, we also show samples of synthetic data and validation set data on the AppWorld and BFCL v3 datasets. It can be seen that the quality of our synthetic data is comparable to the original data, and we can synthesize more difficult data by adjusting the parameters.


\subsection{Performance Comparison}

We compare CuES with representative baselines \citep{grattafiori2024llama, guo2024deepseek, yao2023react} on \textsc{AppWorld}, \textsc{WebShop}, and \textsc{BFCL v3 Multi-Turn Base} under the metrics of \emph{greedy}.

\begin{table}[htbp]
\centering
\caption{Results on \textbf{AppWorld LeaderBoard}.}
\small
\renewcommand\arraystretch{0.8}
\centering
\label{tab:appworld}
\setlength{\tabcolsep}{3.8mm}
\begin{tabular}{l c c c c}
\toprule
\textbf{Model} & \textbf{Think} & \textbf{LLM} & \textbf{Params} & \textbf{greedy} \\
\midrule
FullCodeRefl    &\ding{55}           & GPT-4o           & /    & 33.9\% \\
FullCodeRefl    &\ding{55}         & GPT-4 Turbo      & /    & 25.6\% \\
FullCodeRefl    &\ding{55}        & LLaMA3           & 70B  & 24.4\% \\
FullCodeRefl    &\ding{55}         & DeepSeekCoder    & 33B  & 13.1\% \\
ReAct           &\ding{55}        & GPT-4o           & /  & \textbf{48.8\%} \\
ReAct           &\ding{55}        & GPT-4 Turbo      & /    & 26.8\% \\
ReAct           &\ding{55}        & LLaMA3           & 70B  & 7.1\% \\
ReAct           &\ding{55}        & DeepSeekCoder    & 33B  & 20.8\% \\
PlanExec        &\ding{55}        & GPT-4o           & /    & 44.6\% \\
PlanExec        &\ding{55}        & GPT-4 Turbo      & /    & 32.7\% \\
PlanExec        &\ding{55}        & LLaMA3           & 70B  & 8.9\% \\
PlanExec        &\ding{55}        & DeepSeekCoder    & 33B  & 1.8\% \\
\rowcolor{gray!40}
\textbf{CuES (ours)} &\ding{55}     & Qwen2.5          & 14B  & \underline{45.24\%} \\
\bottomrule
\end{tabular}

\end{table}

\begin{table*}[t]
\small
\renewcommand\arraystretch{0.75}
\centering
\setlength{\tabcolsep}{3.2mm}{
\caption{Results on \textbf{WebShop LeaderBoard}.}
\label{tab:webshop}
\begin{tabular}{l c c c c}
\toprule
\textbf{Model}  & \textbf{Think} & \textbf{LLM} & \textbf{Params} & \textbf{greedy} \\
\midrule
qwen3-14b              &\ding{55}  & Qwen3 & 14B & 30.70\% \\
qwen3-8b               &\ding{55}  & Qwen3 &  8B & 30.01\% \\
qwen3-4b               &\ding{55}  & Qwen3 &  4B & 33.17\% \\
qwen3-235b-a22b        &\ding{55}  & Qwen3 &  235B-A22B & \underline{50.96\%} \\
deepseek-v3            &\ding{55}  & Deepseek &  /  & 26.91\% \\
qwen2.5-3b-instruct    &\ding{55}  & Qwen2.5 &  3B & 23.94\% \\
qwen2.5-7b-instruct    &\ding{55}  & Qwen2.5 &  7B & 22.07\% \\
qwen2.5-14b-instruct   &\ding{55}  & Qwen2.5 & 14B & 23.74\% \\
\rowcolor{gray!40}
\textbf{CuES (ours)} &\ding{55}     & Qwen2.5 & 14B  &\textbf{64.10\%} \\
\bottomrule
\end{tabular}}
\end{table*}

\begin{table*}[t]
\small
\renewcommand\arraystretch{0.75}
\centering
\caption{Results on \textbf{BFCL v3 Multi-Turn Base LeaderBoard}.}
\label{tab:bfclv3}

\begingroup
\setlength{\tabcolsep}{3.5mm}
\begin{tabular}{l c c c c}
\toprule
\textbf{Model} & \textbf{Think} & \textbf{LLM} & \textbf{Params} & \textbf{greedy} \\
\midrule
GPT-5               & \ding{55} & OpenAI           & /                    & 33.5\% \\
GPT-5-mini          & \ding{55} & OpenAI           & /                    & 31.5\% \\
Gemini-2.5-Pro      & \ding{52} & Gemini           & /                    & 35.0\% \\
Gemini-2.5-Flash    & \ding{52} & Gemini           & /                    & 36.0\% \\
GPT-5-nano          & \ding{55} & OpenAI         & /                    & 33.5\% \\
Amazon-Nova-Lite-v1:0 & \ding{55} & Amazon          & /                    & 29.0\% \\
o3                  & \ding{52} & OpenAI o3        & /                    & \underline{44.0\%} \\
DeepSeek-V3         & \ding{55} & DeepSeek         & /                    & 43.5\% \\
GPT-4o-mini         & \ding{55} & OpenAI         & /                    & 43.5\% \\
Amazon-Nova-Pro-v1:0 & \ding{52} & Amazon          & /                    & 42.5\% \\
\rowcolor{gray!40}
\textbf{CuES (ours)}& \ding{52} & Qwen2.5          & 14B                  & \textbf{44.2\%} \\
\bottomrule
\end{tabular}
\endgroup

\end{table*}




\begin{table*}[htbp]
\renewcommand\arraystretch{0.8}
\centering
\setlength{\tabcolsep}{1.3mm}{
\caption{Synthesis-quality ablation with metrics.}
\label{tab:abl_synth_quality_settings_split}
\small
\begin{tabular}{ccccc cccc cccc}
\toprule
\multicolumn{4}{c}{\textbf{Setting}} & & \multicolumn{4}{c}{\textbf{AppWorld}} & \multicolumn{4}{c}{\textbf{WebShop}} \\
\cmidrule(lr){1-4}\cmidrule(lr){6-9}\cmidrule(lr){10-13}
\textbf{min\_conf} & \textbf{batch} & \textbf{max\_steps} & \textbf{Pool} & & \textbf{PR} & \textbf{SR} & \textbf{ED} & \textbf{Step} & \textbf{PR} & \textbf{SR} & \textbf{ED} & \textbf{Step} \\
\midrule
0.7 & 30 & 30 & \ding{55} & & 0.6159 & 0.6477 & 0.0442 & 7.6500 & 0.1667 & 0.6780 & 0.1213 & 7.7812  \\
\midrule
\multicolumn{13}{l}{\emph{Varying min\_confidence (batch=30, max\_steps=30, Req+Pool)}}\\
0.5 & 30 & 30 & \ding{55} & & 0.5868 & 0.6653 & 0.0450 & 7.8979 & 0.1481 & 0.6668 & 0.1143 & 8.6786 \\
0.9 & 30 & 30 & \ding{55} & & 0.6733 & 0.6622 & 0.0467 & 6.8415 & 0.1040 & 0.6772 & 0.1101 & 9.7619 \\
\midrule
\multicolumn{13}{l}{\emph{Varying batch\_size (min\_conf=0.7, max\_steps=30, Req+Pool)}}\\
0.7 & 10 & 30 & \ding{55} & & 0.6359 & \textbf{0.6903} & 0.0369 & 7.1637 & 0.1354 & \textbf{0.7240} & 0.0908  & 5.3191 \\
0.7 & 50 & 30 & \ding{55} & & \textbf{0.7239} & 0.6405 & 0.0385 & 7.7474 & 0.1064 & 0.6864 & 0.1129 & 7.6500 \\
\midrule
\multicolumn{13}{l}{\emph{Varying max\_steps (min\_conf=0.7, batch=30, Req+Pool)}}\\
0.7 & 30 & 20 & \ding{55} & & 0.6439 & 0.6105 & 0.0480 & 7.2000 & 0.1196 & 0.6507 & 0.0899 & 4.9090 \\
0.7 & 30 & 40 & \ding{55} & & 0.6596 & 0.6136 & 0.0438 & \textbf{7.9741} & 0.1475 & 0.6897 & 0.0945 & \textbf{10.5938} \\
\midrule
\multicolumn{13}{l}{\emph{Requirement / Concept Pool toggles (min\_conf=0.7, batch=20, max\_steps=30)}}\\
\rowcolor{gray!40}
0.7 & 30 & 30 & \ding{52} & & 0.6642 & 0.6479 & \textbf{0.0506} & 7.7184 & \textbf{0.2111} & 0.5571 & \textbf{0.3429} & 7.3934 \\
\bottomrule
\end{tabular}}
\end{table*}

In Table \ref{tab:appworld}, \ref{tab:webshop} and \ref{tab:bfclv3}, we compare CUES against recent baselines on AppWorld, WebShop, and BFCL v3 Multi-Turn Base. We report \emph{avg@8} and \emph{greedy} under the no-think and prompt configurations of Qwen2.5 and Qwen3. Since the baseline performance of the Qwen3 series on the BFCL v3 was lower than expected, we report the baseline results of Qwen3 under the thinking mode on the BFCL v3. On AppWorld, CUES attains 45.24\% greedy, which closes to that of the closed-source GPT-4o. The gain is most pronounced on WebShop, where CUES reaches 64.10\% greedy versus 50.96\% for qwen3-235b-a22b. On BFCL v3 Multi-Turn Base, CUES achieves 44.15\% greedy, edging out o3 at 44.0\%. Looking at \ref{tab:scale_perf_compact}, CUES delivers a macro-average of roughly 51.2\% greedy and 50.7\% avg@8 across the three datasets, substantially higher than same-scale non-CUES backbones. These results indicate that environment-grounded synthesis with light top-down steering translates into consistent single-path and sampled execution gains across diverse domains. Notably, even when compared with models that have far more parameters or closed-source model than CUES (e.g., DeepSeek-V3\citep{liu2024deepseek}, GPT-5, o3\citep{achiam2023gpt} and Gemini-2.5 Pro\citep{comanici2025gemini}), CUES still outperforms by a large margin, underscoring the impact of our synthesis rather than raw scale.

We also observe that CUES improves performance even under format shift on BFCL v3 Multi-Turn Base, where evaluation involves multi-turn follow-ups that do not strictly match the form of our synthesized goals. Despite this mismatch, CUES maintains a positive edge on greedy success, suggesting that the synthesized pool emphasizes executable, transferable intents rather than overfitting to any one prompt style. \textbf{Together, these findings support the claim that bottom-up exploration anchored by a concept pool and requirement confirmation can raise executability and coverage in ways that persist across backbones, scales, and interaction protocols.}

\subsection{Hyperparameter Ablations and Distribution Analysis}

For different benchmarks, we observed different changes in synthetic-quality with different settings. We believe that this may be closely related to the tasks. We selected the two benchmarks with the largest differences. 

\paragraph{Approximate metrics.}
Let $X=\{x_i\}$ be TF–IDF+SVD embeddings of \emph{original/target} intents and $Y=\{y_i\}$ embeddings of \emph{synthesized} intents; all vectors are $\ell_2$-normalized so that $\langle x_i,x_j\rangle$ and $\langle y_i,y_j\rangle$ equal cosine similarity. We measure executability, diversity, and relevance by:
\begin{equation}
\label{eq:metrics}
\mathrm{PR}=\frac{|\mathcal{G}_{\mathrm{pass}}|}{|\mathcal{G}|},\qquad
\mathrm{SR@}k=\frac{1}{|Y|}\sum_{i}\frac{1}{k}\sum_{j\in \mathrm{kNN}_Y(i)}\langle y_i,y_j\rangle,\qquad
\mathrm{ED}_{\mathrm{rel}}=\frac{\mathrm{ED}(X,Y)}{\mathbb{E}_{i\neq i'}\|x_i-x_{i'}\|_2}.
\end{equation}
Here $\mathrm{kNN}_Y(i)$ are the $k$ nearest neighbors of $y_i$ in $Y$ under cosine distance. The energy distance is
\begin{equation}
\label{eq:ed_def}
\mathrm{ED}(X,Y)=\frac{2}{|X||Y|}\sum_{i=1}^{|X|}\sum_{j=1}^{|Y|}\|x_i-y_j\|_2
-\frac{1}{|X|^2}\sum_{i=1}^{|X|}\sum_{i'=1}^{|X|}\|x_i-x_{i'}\|_2
-\frac{1}{|Y|^2}\sum_{j=1}^{|Y|}\sum_{j'=1}^{|Y|}\|y_j-y_{j'}\|_2.
\end{equation}
Intuitively, higher $\mathrm{PR}$ indicates valid, verifiable goals; lower $\mathrm{SR@}k$ indicates non-redundant coverage; lower $\mathrm{ED}_{\mathrm{rel}}$ indicates distributional alignment to the target.

Ref to \ref{tab:abl_synth_quality_settings_split}, We assess synthesis quality with sentence embedding. It is worth noting that the step in the table here refers to the step generated by Quality Control re-executes. Executability is measured by Pass Rate, diversity by Self-Redundancy over $k$ nearest neighbors, and distributional alignment by a relative energy distance between original and synthesized intent embeddings. 

\textbf{On \textsc{AppWorld}, higher confidence improves executability with small costs to diversity and alignment.} Raising the threshold from 0.7 to 0.9 lifts PR from 0.6159 to 0.6733, while SR and ED change only slightly and the average step count falls from 7.65 to 6.84. Larger batches are effective: batch 50 reaches PR 0.7239 with controlled SR 0.6405 and ED 0.0385, outperforming batch 10 on PR while keeping alignment competitive. Extending the rollout depth from 20 to 40 steps increases PR to 0.6596 and reduces ED to 0.0438.

\textbf{\textsc{WebShop} favors tighter alignment through smaller batches and shorter rollouts, while still gaining executability when needed.} With min\_conf at 0.7, batch 10 improves ED to 0.0908 and reduces steps to 5.32 compared with batch 50, though PR is lower at 0.1354 versus 0.1064 for batch 50 the alignment gain is clear. Shorter rollouts also help alignment: max\_steps 20 attains ED 0.0899 and Step 4.91, whereas max\_steps 40 raises PR to 0.1475 with a mild ED increase to 0.0945 and a higher step cost. Enabling the concept pool strongly boosts PR from 0.1667 to 0.2111 and lowers SR from 0.6780 to 0.5571, yet shifts ED upward to 0.3429, reflecting targeted exploration toward salient subspaces. 

These conclusions indicate that the design of CuES conforms to the Sec.\ref{sec:formu} guiding $F_{task}$. CuES performs well in terms of Executability, Diversity, and Relevance under different benchmarks and hyperparameters.

\begin{figure}[t]
    \centering
    \begin{minipage}[t]{0.48\linewidth}
        \centering
        \includegraphics[width=\linewidth]{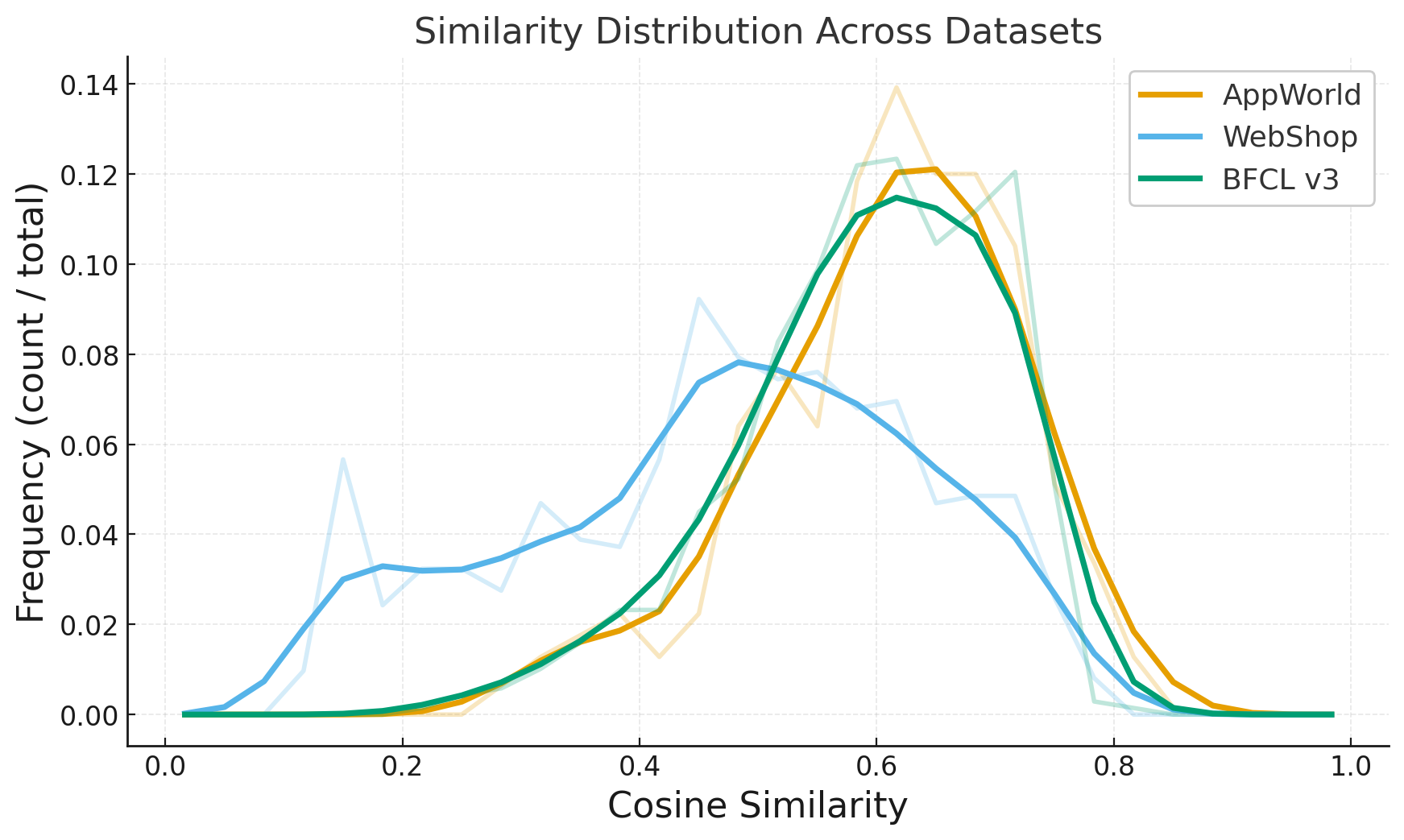}
        \vspace{-2mm}
        \subcaption{Similarity distribution across datasets.}
    \end{minipage}
    \hfill
    \begin{minipage}[t]{0.48\linewidth}
        \centering
        \includegraphics[width=\linewidth]{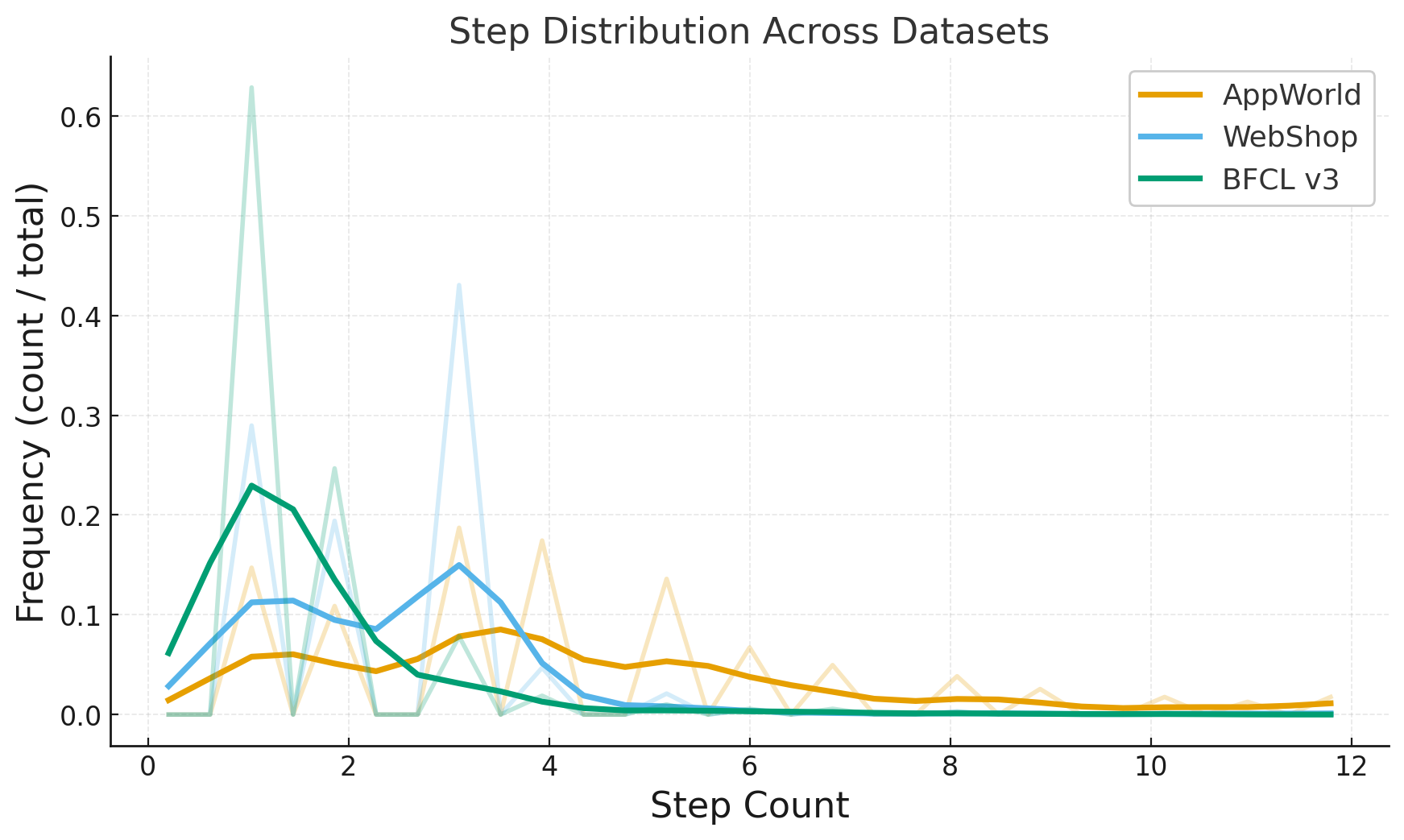}
        \vspace{-2mm}
        \subcaption{Step of guideline distribution across datasets.}
    \end{minipage}
    \vspace{-2mm}
    \caption{
    \textbf{Distributions of similarity and step count across datasets.}
    (\textbf{Left}) The similarity distribution reveals that AppWorld and BFCL v3 exhibit higher average cosine similarity compared to WebShop, suggesting more internally coherent samples. 
    (\textbf{Right}) The step distribution indicates distinct task complexity patterns across environments, with WebShop containing longer interaction chains on average. 
    Together, these distributions highlight the diversity and varying difficulty of the evaluation environments used in this study.
    }
    \label{fig:distribution_summary}
\end{figure}

\subsection{Visualization of Synthesized and Original goals}

Figure~\ref{fig:dist_grid} shows t-SNE projections of sentence embeddings for original and CUES-synthesized goals on AppWorld, WebShop, and BFCL v3. In AppWorld, the synthesized cloud envelops the sparse original points with nearly coincident centroids and small drift, indicating that CUES expands coverage around existing intent modes while remaining aligned. BFCL v3 exhibits the strongest agreement, with tightly interleaved clouds and very low relative energy distance, reflecting close intent matching. WebShop presents a deliberate shift: the synthesized cloud moves below the original cluster and ED\_rel increases, while self-redundancy decreases, revealing broader variety than the original set, which was concentrated in a narrow region.

\textbf{We complement the plots with concise qualitative samples. On both benchmark, the synthetic sample showed better quality and diversity than the original sample. } Ref to Fig.\ref{fig:samples}, in AppWorld, CUES surfaces executable workflows in the Spotify domain (login, enumerate genres, filter artists, follow), making affordances explicit and explaining the low drift and high pass rate. In BFCL v3, synthesized tasks articulate multi-system checks (vehicle states, controls, and brief terminal interactions), demonstrating precise tool grounding consistent with the tight embedding overlap. 

Figure~\ref{fig:distribution_summary} visualizes the distribution of maximum cosine similarity and step between each synthesized goal and the closest seed string. CuES preserves proximity to seeds in AppWorld and BFCL while expanding into novel regions; in WebShop it intentionally broadens coverage beyond a narrow seed cluster. And the overall distribution maintains a certain distance from the validation set.

\section{Conclusion}

We introduce CUES, a curiosity-guided, environment-grounded synthesis framework that unifies bottom-up exploration with top-down guidance to generate high-quality, executable agent data in goal-free settings. Unlike prior methods that either decouple tasks from environment dynamics or drift under unconstrained exploration, CUES integrates lightweight top-down guidance—via requirement confirmation and concept pools—with bottom-up discovery powered by environment memory and intrinsic curiosity, \textbf{yielding diverse yet solvable trajectories while sharply reducing ineffective exploration.} Experiments on AppWorld, WebShop, and BFCL v3 show that CUES-synthesized data not only match but surpass original datasets, outperform strong baselines by an average of over 30 points on avg@8 and greedy metrics, and remain effective even against models with far larger parameter scales. These results demonstrate that curiosity-driven bottom-up synthesis, when coupled with minimal but strategic top-down control, tying executability, diversity, and distributional alignment to controllable hyperparameters, provides a scalable recipe for robust agent training in real environments. We view this top-down–meets–bottom-up design as a step toward reliable agent training corpora that adapt across domains and interaction protocols, and as a foundation for future on-policy synthesis and environment-specific reward modeling.

\newpage

\bibliography{colm2025_conference}
\bibliographystyle{colm2025_conference}

\newpage

\end{document}